\documentclass{article}

\usepackage[preprint]{neurips_2026}

\usepackage[utf8]{inputenc} 
\usepackage[T1]{fontenc}    
\usepackage{hyperref}       
\usepackage{url}            
\usepackage{booktabs}       
\usepackage{amsfonts}       
\usepackage{nicefrac}       
\usepackage{microtype}      
\usepackage{xcolor}         
\usepackage{amsmath}
\usepackage{graphicx}
\usepackage{natbib}
\usepackage{multirow}
\usepackage{subcaption}
\usepackage{amsthm}
\usepackage{array}
\usepackage{tabularx}
\usepackage{makecell}
\usepackage{array}  

\definecolor{col}{HTML}{598BE7}
\hypersetup{
  colorlinks=true,
  linkcolor=col,   
  citecolor=col,    
  urlcolor=col    
}

\definecolor{easybg}{HTML}{FFE0B2}   
\definecolor{medbg}{HTML}{FFF2B2}    
\definecolor{hardbg}{HTML}{D7E7FF}   
\definecolor{hardblue}{HTML}{2F5BBF}

\newcommand{\easytok}[1]{%
  \begingroup\setlength{\fboxsep}{1.6pt}%
  \colorbox{easybg}{\strut #1}%
  \endgroup%
}
\newcommand{\medtok}[1]{%
  \begingroup\setlength{\fboxsep}{1.6pt}%
  \colorbox{medbg}{\strut #1}%
  \endgroup%
}
\newcommand{\hardtok}[1]{%
  \begingroup\setlength{\fboxsep}{1.6pt}%
  \colorbox{hardbg}{\strut #1}%
  \endgroup%
}

\newcolumntype{R}[1]{>{\raggedleft\arraybackslash}p{#1}}
\newcolumntype{L}[1]{>{\raggedright\arraybackslash}p{#1}}

\title{\!When Are Experts Misrouted? Counterfactual Routing\!\!\\Analysis in Mixture-of-Experts Language Models}

\author{%
Youngsik Yoon\textsuperscript{1} \quad
  Siwei Wang\textsuperscript{3} \quad
  Wei Chen\textsuperscript{3} \quad
  Jungseul Ok\textsuperscript{1,2}\thanks{Corresponding author.} \\
  \textsuperscript{1}Department of Computer Science and Engineering, POSTECH, South Korea \\
  \textsuperscript{2}Graduate School of Artificial Intelligence, POSTECH, South Korea \\
  \textsuperscript{3}Microsoft Research Asia, Beijing, China \\
  \texttt{\{ysyoon97, jungseul.ok\}@postech.ac.kr, \{siweiwang, weic\}@microsoft.com}
}

\begin{document}

\maketitle

\begin{abstract}
Mixture-of-Experts (MoE) language models route each token to a small subset of experts, but whether the routes selected by a trained top-$k$ router are good ones is rarely evaluated directly. 
Holding the model fixed, we compare each standard route against sampled equal-compute alternatives for the same token and score each by the next-token probability it assigns to the realized token in a verified reasoning trajectory. 
The result is sharply token-conditional: the standard router is well-aligned with route utility on confident tokens but uninformative on the fragile tokens that drive hard reasoning, where lower-loss equal-compute routes consistently exist inside the frozen model but are not selected. 
The same pattern holds across Qwen3-30B-A3B, GPT-OSS-20B, DeepSeek-V2-Lite, and OLMoE-1B-7B, and follows structurally from how standard top-$k$ training evaluates routing decisions: the language modeling loss scores only the executed route, and load balancing depends only on aggregate routing statistics.
A minimal router-only update to the final-layer router, leaving every expert and every other router frozen, is sufficient to shift pass@K on AIME 2024+2025 and HMMT 2025 for both Qwen3-30B-A3B and GPT-OSS-20B, suggesting that at least part of the failure reflects router-reachable misallocation rather than expert capacity alone.
\end{abstract}
\section{Introduction}
\label{sec:introduction}

Mixture-of-Experts (MoE) language models provide an efficient way to scale model capacity by activating only a small subset of parameters for each token. By routing each token to a few experts rather than executing the full model, sparse MoE models can scale total parameters while keeping per-token compute roughly fixed~\citep{shazeer2017outrageously, lepikhin2020gshard, fedus2022switch, du2022glam}. This design has become a common choice in recent frontier and open-weight language models, including Mixtral~\citep{jiang2024mixtral}, DeepSeek-V3/R1~\citep{liu2024deepseek3, guo2025deepseekr1}, Qwen3~\citep{yang2025qwen3}, and GPT-OSS~\citep{agarwal2025gpt}.

The component that decides which experts process each token is the router, 
typically a single learned linear projection followed by top-$k$ selection 
over experts. 
Compared with the attention and expert components, the router 
has received relatively little attention: standard MoE pretraining treats routing as a 
balanced-allocation problem solved by load-balancing \citep{shazeer2017outrageously, fedus2022switch} and stability regularizers \citep{zoph2022st}, with downstream performance taken as sufficient 
evidence of good routing. 
Analyses of trained routers report that routing 
stabilizes early and exhibits structured specialization 
\citep{dai2022stablemoe, xue2024openmoe}, properties read as the router 
doing its job, and the apparent adequacy of these routers is rarely 
questioned in subsequent analysis.

\begin{figure}[t]
\centering
\small

\begin{minipage}[t]{0.48\linewidth}
\centering

\vspace{0.4em}

\fbox{%
\begin{minipage}{0.94\linewidth}
\footnotesize
\setlength{\parskip}{0.45em}

\easytok{\$}
\easytok{$\lceil$}
\easytok{64}
\easytok{/}
\easytok{2}
\easytok{$\rceil$}
\easytok{=}
\easytok{32}
\easytok{\$}
\easytok{,}
\easytok{\$}
\easytok{$\lfloor$}
\easytok{64}
\easytok{/}
\easytok{2}
\easytok{$\rfloor$}
\easytok{=}
\easytok{32}
\easytok{\$}
\easytok{.}
\easytok{So}
\hardtok{all}
\easytok{8}
\easytok{\texttimes}
\easytok{8}
\easytok{squares}
\hardtok{have}
\medtok{exactly}
\easytok{32}
\easytok{black}
\easytok{squares}
\easytok{\textrightarrow}
\easytok{\textgreater=}
\easytok{5}
\easytok{\textrightarrow}
\medtok{qualifies}
\easytok{.}
\medtok{Only}
\medtok{one}
\easytok{such}
\easytok{square}
\easytok{:}
\easytok{the}
\easytok{whole}
\easytok{board}
\easytok{\textrightarrow}
\easytok{1}
\easytok{valid}
\easytok{square}
\\
...
\vspace{-0.2cm}

\begin{center}
\setlength{\fboxsep}{1.1pt}
\colorbox{easybg}{\strut \phantom{xx}} \; Confident
\quad
\colorbox{medbg}{\strut \phantom{xx}} \; Ambiguous
\quad
\colorbox{hardbg}{\strut \phantom{xx}} \; Fragile
\end{center}

\end{minipage}
}

\subcaption[]{Confidence along a reasoning trace.}

\end{minipage}
\hfill
\begin{minipage}[t]{0.5\linewidth}
\centering

\vspace{0.8em}

\small
\setlength{\tabcolsep}{3.5pt}
\begin{tabular}{lrrr}
\toprule
\multirow{2}{*}{Confidence}
& \multicolumn{1}{c}{Confident}
& \multicolumn{1}{c}{Ambiguous}
& \multicolumn{1}{c}{Fragile} \\
& \multicolumn{1}{c}{\tiny $\bar p > 0.9$}
& \multicolumn{1}{c}{\tiny $0.9 \ge \bar p > 0.5$} 
& \multicolumn{1}{c}{\tiny $0.5 \ge \bar p$}
\\
\midrule
Tokens (\%)
& 78.7
& 14.3
& 6.9 \\

Top-1 (\%)
& 51.9
& 1.5
& 0.8 \\

$p_{\rm std}$ (\%)
& 99.6
& 77.9
& 40.2 \\

$p_{\rm best}$ (\%)
& 99.8
& 88.9
& 60.6 \\

Gap (pp)
& 0.2
& 11.0
& 20.4 \\
\bottomrule
\end{tabular}
\subcaption[]{Routing gaps by confidence bin.}
\end{minipage}
\caption{
\textbf{Counterfactual routing statistics by confidence bin.}
Tokens are grouped by $\bar p$, the mean realized-token probability over sampled equal-compute alternative routes: Confident ($\bar p>0.9$), Ambiguous ($0.5<\bar p\le0.9$), and Fragile ($\bar p\le0.5$). 
Top-1 reports how often the standard top-$k$ route is best among evaluated routes. 
$p_{\mathrm{std}}$ and $p_{\mathrm{best}}$ denote realized-token probabilities under the standard and best evaluated routes, and Gap is their difference. 
Results are for Qwen3-30B-A3B at the final MoE layer on verified MATH Level-5 trajectories.
}
\label{fig:easy-medium-hard-routing}
\vspace{-0.5cm}
\end{figure}

We ask whether, in MoE language models, the routing decisions actually made by the standard top-$k$ router are good ones. 
Holding the model fixed, we compare each standard top-$k$ route against sampled equal-compute alternatives of the same size and score each route by the next-token probability it assigns to the realized token in a verified reasoning trajectory.
This question is especially important at low-confidence reasoning tokens. 
Prior work shows that reasoning success can depend disproportionately on a small subset of pivotal, critical, or high-entropy tokens \citep{abdin2024phi,lin2024critical,wang2025beyond}. 
We therefore group tokens by counterfactual confidence: the average probability assigned to the realized next token by sampled alternative routes.

Figure~\ref{fig:easy-medium-hard-routing} shows the core pattern. 
On Confident tokens, the standard route is often the best evaluated route and closely matches the best sampled route in next-token probability. 
On Fragile tokens, this alignment collapses: the standard route is best on only 0.8\% of tokens, while the best sampled equal-compute route improves next-token probability by 20.4 percentage points. 
Thus the failure is not uniform routing noise; it concentrates on the low-confidence tokens where route choice is most consequential. 
The same pattern holds across layers, benchmarks, and MoE model families.

The misalignment we observe is consistent with a structural property of standard top-$k$ MoE training: the language modeling loss only evaluates the route that was executed, while load balancing and other auxiliary losses depend only on aggregate router-side statistics, not on the loss that would have resulted from a different route on the same token.

We then test whether this routing failure is partially reachable inside the frozen trained model. Updating only the final-layer router with counterfactual route preferences, while freezing all experts and earlier-layer routers (less than 0.001\% of model parameters), is sufficient to shift pass@K on AIME 2024+2025 and HMMT 2025 for both Qwen3-30B-A3B and GPT-OSS-20B. That such a small router-only update shifts downstream performance supports the reading that some high-loss predictions on hard tokens reflect a routing failure rather than a pure capacity limit.

In summary, this paper provides a counterfactual analysis of routing quality in trained MoE language models, showing that the standard router can be well-aligned on confident tokens but systematically misaligned on harder ones, and that this misalignment is partially reachable through a minimal router-only update. Our results suggest that routing quality should be considered as a first-class concern in MoE training, not just as an implementation detail of sparsity.
\section{Preliminaries}
\label{sec:preliminaries}

A sparse MoE layer replaces a single dense feed-forward block with $N$ parallel expert modules $\{E_i\}_{i=1}^{N}$, each $E_i: \mathbb{R}^{d} \to \mathbb{R}^{d}$, and routes each token to a small subset of them. Let $x_t \in \mathbb{R}^{d}$ denote the hidden representation of token position $t$ entering the layer, where $d$ is the model width.

The routing decision is made by a learned linear router with weights $W_r \in \mathbb{R}^{N \times d}$ and bias $b_r \in \mathbb{R}^{N}$, which produces a vector of expert scores $s_t = W_r x_t + b_r \in \mathbb{R}^{N}$, where $s_{ti}$ is the score assigned to expert $i$ for token $t$. This single linear projection is the standard router parameterization in modern open-weight MoE language models (Table~\ref{tab:models}). The standard top-$k$ route is the index set of the $k$ highest-scoring experts,
\begin{equation}
S_t^{\mathrm{std}} = \mathrm{TopK}(s_t, k) \subseteq \{1, \ldots, N\}, \qquad |S_t^{\mathrm{std}}| = k,
\end{equation}
where $k \ll N$ in practice, with $k=8$ out of $N=128$ in Qwen3-30B-A3B as a typical example.

Only the selected experts are executed. Their outputs are combined into the layer output by softmax-normalizing the router scores within $S_t^{\mathrm{std}}$,
\begin{equation}
h_t^{\mathrm{std}} = \sum_{i \in S_t^{\mathrm{std}}} g_{ti}^{\mathrm{std}} \, E_i(x_t), \qquad g_{ti}^{\mathrm{std}} = \frac{\exp(s_{ti})}{\sum_{j \in S_t^{\mathrm{std}}} \exp(s_{tj})},
\end{equation}
so that the gate weights $g_{ti}^{\mathrm{std}}$ form a distribution over $S_t^{\mathrm{std}}$. Unselected experts contribute neither their outputs nor their scores to $h_t^{\mathrm{std}}$. The next-token cross-entropy at position $t$, $C_t = -\log p_\theta(y_t \mid x, y_{<t})$ where $y_t$ is the realized next token and $\theta$ collects all model parameters, is therefore a function of the standard route $S_t^{\mathrm{std}}$ alone when $\theta$ is fixed.
\section{Evaluating Routing Quality}
\label{sec:analysis}

Given a trained MoE language model, we ask whether its router selects expert sets that are really useful for the tokens it routes.
We answer this by comparing the standard top-$k$ route against sampled alternatives for the same token.
Section~\ref{sec:protocol} describes the analysis protocol, including how we sample alternative routes and measure their utility. Section~\ref{sec:main_finding} reports our main finding: the standard router is well-aligned with route utility on confident tokens, but this alignment collapses on harder tokens, where the standard route is rarely among the lowest-loss candidates. Section~\ref{sec:consistency} shows that this pattern persists across MoE layers, evaluation domains, and model families.

\begin{table}[t]
\centering
\caption{MoE language models used in our experiments. The expert column shows total / active experts per layer, with shared experts (always activated) listed after a `+'.}
\label{tab:models}
\small
\begin{tabular}{lccc}
\toprule
Model & Total / Active Params & Layers & $N$ / $k$ \\
\midrule
Qwen3-30B-A3B~\citep{yang2025qwen3} & 30.5B / 3.3B & 48 & 128 / 8 \\
GPT-OSS-20B~\citep{agarwal2025gpt} & 21B / 3.6B & 24 & 32 / 4 \\
DeepSeek-V2-Lite~\citep{liu2024deepseek2} & 16B / 2.4B & 27 & 64 / 6 + 2 \\
OLMoE-1B-7B~\citep{muennighoff2024olmoe} & 7B / 1.3B & 16 & 64 / 8 \\
\bottomrule
\end{tabular}
\vspace{-0.3cm}
\end{table}

\subsection{Analysis Protocol}
\label{sec:protocol}

\paragraph{Setup.}
We analyze four open-weight MoE language models (Table~\ref{tab:models}) on verified self-generated reasoning trajectories. 
Our primary analysis uses Qwen3-30B-A3B on MATH Level-5~\citep{hendrycks2021measuring}; we extend to the other models and to AIME, HMMT, and GPQA-Diamond~\citep{rein2023gpqa} in Section~\ref{sec:consistency}. 
For each problem we sample candidate solutions and retain one verified-correct trajectory $y_{1:T}$ generated under prompt $x$, restricted to assistant-response tokens.

\paragraph{Sampled alternative routes.}
For token $t$ at MoE layer $\ell$, let $S_{t,\ell}^{\mathrm{std}}$ denote the standard top-$k$ route. 
We form a candidate expert pool $P_{t,\ell}$ consisting of the $m=32$ highest-scoring routed experts under the trained router at that token and layer. 
We compare the standard route against $G=32$ equal-compute alternatives drawn within this pool via Gumbel-top-$k$ sampling: for each $g = 1, \ldots, G$, we draw independent Gumbel noise $\boldsymbol{\epsilon}^{(g)} \sim \mathrm{Gumbel}(0, 1)^{m}$ and take $S_{t,\ell}^{(g)} = \mathrm{TopK}_{i\in P_{t,\ell}}(s_{t,\ell,i} + \boldsymbol{\epsilon}^{(g)}, \, k)$.
This yields a candidate set $\mathcal{A}_{t,\ell} = \{S_{t,\ell}^{\mathrm{std}}, S_{t,\ell}^{(1)}, \ldots, S_{t,\ell}^{(G)}\}$ of $G + 1 = 33$ routes.
For each $S \in \mathcal{A}_{t,\ell}$, we intervene at $(t, \ell)$ by replacing the standard route with $S$, recompute the gate weights, and run the rest of the model forward with all other parameters fixed.

\paragraph{Routing-quality metrics.}
Recent work has shown that a small subset of pivotal tokens disproportionately determines reasoning success~\citep{abdin2024phi, lin2024critical}, and that local signals such as next-token entropy can identify them~\citep{wang2025beyond}. We use trajectories that are verified correct, so each realized next token lies on a successful path; the probability a route assigns to this token is a necessary condition for routing well at $t$, and serves as a local proxy for routing quality.
Based on this successful path, we report the standard-route probability $p_{t,\ell}(S_{t,\ell}^{\mathrm{std}})$, the best-route probability $\max_S p_{t,\ell}(S)$, their gap, and the Top-$K$ rates for $K \in \{1, 5, 10\}$, defined as the fraction of tokens at which the standard route ranks within the top $K$ of the $33$ candidates.
We stratify tokens by route-averaged probability $\bar{p}_{t,\ell} = \frac{1}{G} \sum_{S \in \mathcal{A}_{t,\ell}\setminus{\{S_{t,\ell}^{std}}\}} p_{t,\ell}(S)$ into 
{\it Confident} ($(0.9, 1]$), {\it Ambiguous} ($(0.5, 0.9]$), and {\it Fragile} ($[0,0.5]$).

\begin{figure}[t]
\centering
\begin{subfigure}[b]{0.48\textwidth}
    \centering
    \includegraphics[width=\textwidth]{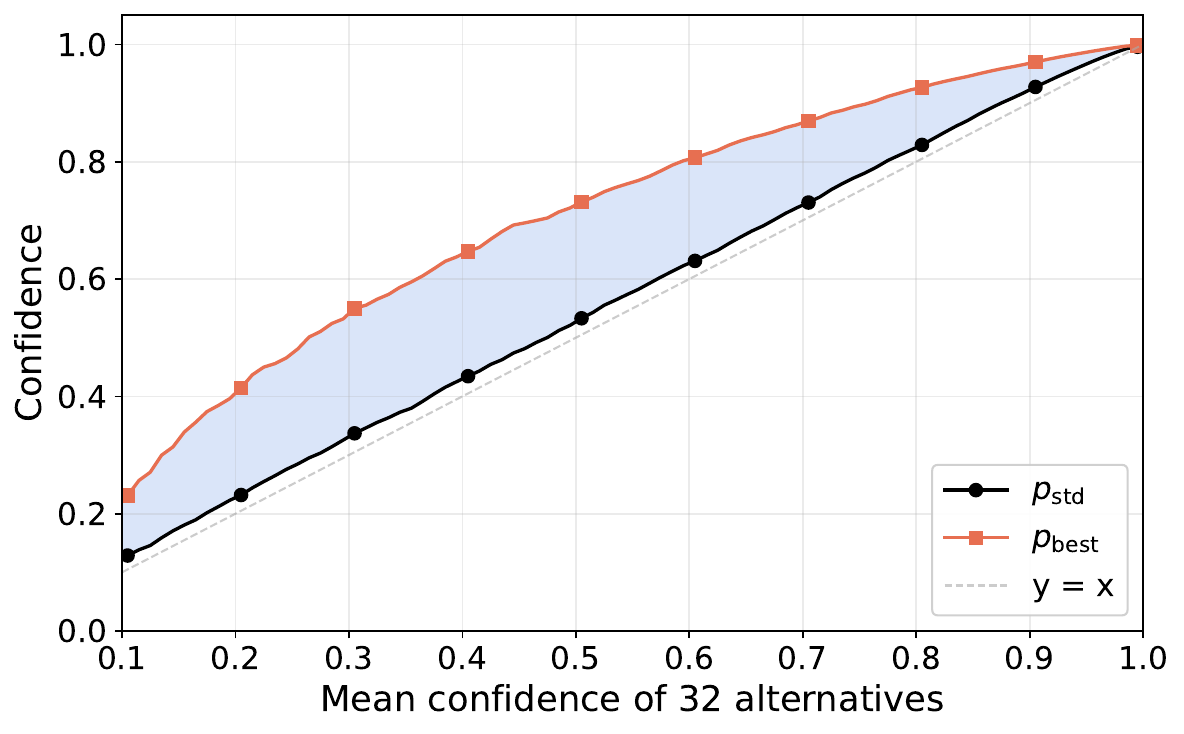}
    \caption{Standard vs. best route probability}
    \label{fig:routing_gap}
\end{subfigure}
\hfill
\begin{subfigure}[b]{0.48\textwidth}
    \centering
    \includegraphics[width=\textwidth]{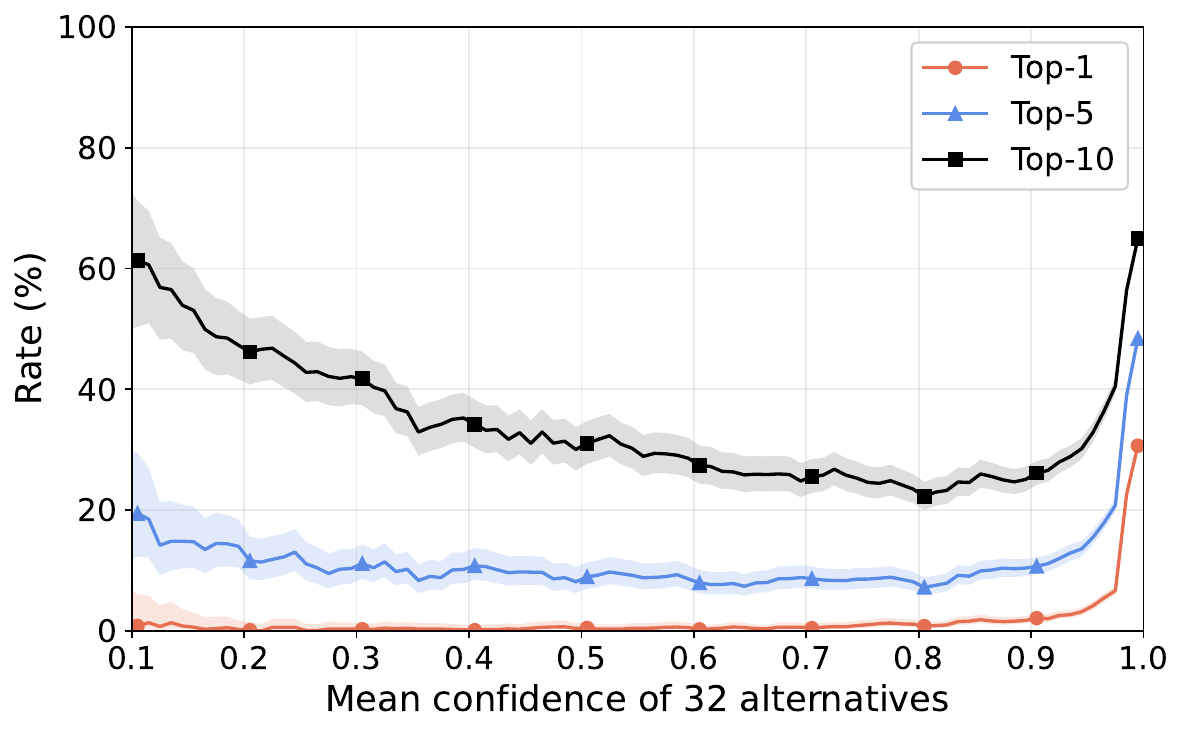}
    \caption{Standard route Top-$K$ performance}
    \label{fig:topk_rates}
\end{subfigure}

\caption{\textbf{Routing quality degrades sharply on low-confidence tokens.} 
(a) Standard route probability vs. best route probability across token difficulty, showing gap as tokens become harder. 
(b) Standard route performance measured by Top-$K$ rates.}
\label{fig:routing_analysis}
\vspace{-0.5cm}
\end{figure}

\subsection{Routing Quality Degrades with Token Difficulty}
\label{sec:main_finding}

We apply the protocol of Section~\ref{sec:protocol} to Qwen3-30B-A3B at the final MoE layer on verified MATH Level-5 trajectories, comparing each standard route against $G = 32$ sampled equal-compute alternatives.
The resulting statistics, stratified by token difficulty, are summarized in Figure~\ref{fig:routing_analysis}.

In Figure~\ref{fig:routing_gap}, the standard-route probability tracks the route-averaged probability $\bar p_{t,\ell}$ closely across the full range. 
On Confident tokens, this is not a demanding test of routing quality: all evaluated routes already assign very high probability to the realized next token, so the difference between the standard route and the best evaluated route is negligible. 
In this regime, routing choice has little observable consequence.
The important behavior appears once $\bar p_{t,\ell}$ leaves the Confident range. 
As tokens become Ambiguous or Fragile, the best evaluated route remains substantially above the standard route, showing that a same-compute alternative inside the frozen model assigns much higher probability to the successful continuation. 
Thus the failure is not uniform across tokens; it appears precisely where route choice begins to matter.

Figure~\ref{fig:topk_rates} shows the same pattern in rank space. 
High Top-$K$ rates on Confident tokens are expected because many routes already perform nearly perfectly. 
Outside this high-probability regime, however, the standard route’s rank collapses: Top-1, Top-5, and Top-10 rates drop sharply and remain near the floor. 
The router is therefore not merely missing small gains on easy tokens; it becomes uninformative on the tokens where alternative routes differ meaningfully in realized-token probability.

\subsection{Consistency across Layers, Domains, and Models}
\label{sec:consistency}

The pattern in Section~\ref{sec:main_finding} is not specific to the final MoE layer of Qwen3-30B-A3B on MATH Level-5.
We repeat the analysis along three axes: across other MoE layers within Qwen3-30B-A3B, across other reasoning and knowledge benchmarks, and across other MoE language models.
The qualitative shape is preserved (Tables~\ref{tab:layerwise_routing}--\ref{tab:cross_model_routing}); only the magnitude varies, in interpretable ways.

\paragraph{Layers.}
We repeat the analysis at an early (L$0$), middle (L$24$), and final (L$47$) MoE layer of Qwen3-30B-A3B (Table~\ref{tab:layerwise_routing}).
On Fragile tokens, the Top-$1$ rate is at or near the uniform baseline at all three layers, so the misalignment is not specific to the final layer.
The Fragile-token gap, however, grows monotonically with depth ($11.4$, $12.8$, $20.4$), which is consistent with the fact that an error at the final MoE layer feeds directly into the next-token distribution with no further routing left to correct it.
We accordingly focus on the final layer for the rest of the analysis and for the recovery experiment in Section~\ref{sec:experiments}.

\begin{table}[!t]
\centering
\caption{\textbf{Layer-wise routing quality.}
L0, L24, L47 refer to early, middle, and final MoE layers respectively.
Tokens are binned by route-averaged next-token probability.
Gap shows probability improvement from standard to best route.
All values are in percent.
}
\label{tab:layerwise_routing}
\footnotesize
\setlength{\tabcolsep}{4pt}
\begin{tabular}{lR{1.08cm}R{1.08cm}R{1.08cm}R{1.08cm}R{1.08cm}R{1.08cm}R{1.08cm}R{1.08cm}R{1.08cm}}
\toprule
\multirow{2}{*}{Layer} & \multicolumn{3}{c}{Confident} & \multicolumn{3}{c}{Ambiguous} & \multicolumn{3}{c}{Fragile} \\
 \cmidrule(lr){2-4} \cmidrule(lr){5-7} \cmidrule(lr){8-10}
 & L0 & L24 & L47 & L0 & L24 & L47 & L0 & L24 & L47 \\
\midrule
Tokens (\%) & 75.9 & 72.8 & 78.7 & 16.7 & 18.7 & 14.3 & 7.4 & 8.5 & 6.9 \\
Top-1 (\%) & 34.8 & 21.9 & 51.9 & 4.1 & 2.6 & 1.5 & 4.1 & 2.9 & 0.8 \\
Top-5 (\%) & 48.7 & 38.5 & 68.8 & 19.3 & 16.5 & 9.6 & 21.3 & 19.0 & 10.5 \\
Top-10 (\%) & 61.4 & 55.4 & 80.5 & 36.8 & 36.0 & 26.5 & 40.5 & 40.7 & 36.7 \\
$p_{\mathrm{std}}$ (\%)& 99.4 & 99.2 & 99.6 & 74.9 & 74.6 & 77.9 & 36.9 & 37.0 & 40.2 \\
$p_{\mathrm{best}}$ (\%)& 99.7 & 99.6 & 99.8 & 83.4 & 83.9 & 88.9 & 48.3 & 49.8 & 60.6 \\
Gap (pp)& 0.3 & 0.4 & 0.2 & 8.5 & 9.3 & 11.0 & 11.4 & 12.8 & 20.4 \\
\bottomrule
\end{tabular}
\vspace{-0.2cm}
\end{table}

\begin{table}[!t]
\centering
\caption{\textbf{Routing quality across reasoning and knowledge benchmarks.} AIME pools 2024 and 2025 (10 trajectories), HMMT 2025 (10 trajectories), and GPQA-Diamond (100 trajectories). Conf., Amb., and Frag. denote the 
Confident, Ambiguous, and Fragile bins, respectively.}
\label{tab:detailed_domain_routing}
\footnotesize
\setlength{\tabcolsep}{4pt}
\begin{tabular}{lR{1.08cm}R{1.08cm}R{1.08cm}R{1.08cm}R{1.08cm}R{1.08cm}R{1.08cm}R{1.08cm}R{1.08cm}}
\toprule
\multirow{2}{*}{Benchmark} & \multicolumn{3}{c}{AIME} & \multicolumn{3}{c}{HMMT} & \multicolumn{3}{c}{GPQA} \\
\cmidrule(lr){2-4} \cmidrule(lr){5-7} \cmidrule(lr){8-10}
 & Conf. & Amb. & Frag. & Conf. & Amb. & Frag. & Conf. & Amb. & Frag. \\
\midrule
Tokens (\%)& 71.6 & 17.9 & 10.5 & 74.6 & 16.3 & 9.2 & 78.2 & 16.4 & 5.5 \\
Top-1 (\%)& 40.8 & 1.2 & 0.6 & 44.4 & 1.3 & 0.7 & 40.3 & 2.6 & 1.6 \\
Top-5 (\%)& 63.2 & 10.2 & 11.9 & 63.6 & 8.7 & 10.4 & 58.9 & 12.1 & 14.0 \\
Top-10 (\%)& 75.8 & 26.8 & 37.5 & 76.3 & 24.3 & 37.0 & 75.4 & 31.6 & 44.6 \\
$p_{\mathrm{std}}$ (\%)& 99.5 & 77.6 & 37.4 & 99.5 & 76.9 & 37.5 & 99.6 & 79.4 & 50.0 \\
$p_{\mathrm{best}}$ (\%)& 99.8 & 88.2 & 56.8 & 99.8 & 87.9 & 56.5 & 99.8 & 90.5 & 71.2 \\
Gap (pp)& 0.3 & 10.6 & 19.4 & 0.3 & 11.0 & 19.0 & 0.2 & 11.1 & 21.2 \\
\bottomrule
\end{tabular}
\vspace{-0.2cm}
\end{table}

\begin{table}[!t]
\centering
\caption{\textbf{Routing quality across MoE language models.} Each model is evaluated with its native top-$k$ configuration.}
\label{tab:cross_model_routing}
\footnotesize
\setlength{\tabcolsep}{4pt}
\begin{tabular}{lR{1.08cm}R{1.08cm}R{1.08cm}R{1.08cm}R{1.08cm}R{1.08cm}R{1.08cm}R{1.08cm}R{1.08cm}}
\toprule
\multirow{2}{*}{Model} & \multicolumn{3}{c}{GPT-OSS} & \multicolumn{3}{c}{DeepSeek} & \multicolumn{3}{c}{OLMoE} \\
\cmidrule(lr){2-4} \cmidrule(lr){5-7} \cmidrule(lr){8-10}
 & Conf. & Amb. & Frag. & Conf. & Amb. & Frag. & Conf. & Amb. & Frag. \\
\midrule
Tokens (\%)& 71.8 & 15.4 & 12.8 & 67.0 & 21.1 & 11.9 & 52.5 & 32.7 & 14.8 \\
Top-1 (\%)& 13.9 & 2.9 & 1.8 & 4.6 & 1.1 & 0.6 & 43.0 & 23.9 & 6.4 \\
Top-5 (\%)& 46.0 & 22.3 & 23.7 & 18.0 & 5.8 & 5.1 & 84.1 & 58.9 & 27.8 \\
Top-10 (\%)& 65.9 & 42.1 & 44.8 & 46.2 & 23.2 & 26.2 & 95.8 & 80.9 & 59.0 \\
$p_{\mathrm{std}}$ (\%)& 99.2 & 75.9 & 26.2 & 99.5 & 85.6 & 42.6 & 99.9 & 95.3 & 57.3 \\
$p_{\mathrm{best}}$ (\%)& 99.6 & 85.9 & 38.1 & 99.8 & 94.3 & 67.5 & 99.9 & 98.2 & 78.1 \\
Gap (pp)& 0.4 & 9.9 & 11.8 & 0.3 & 8.7 & 24.9 & 0.1 & 2.9 & 20.8 \\
\bottomrule
\end{tabular}
\vspace{-0.2cm}
\end{table}

\paragraph{Domains.}
We extend the analysis to AIME, HMMT, and GPQA-Diamond on Qwen3-30B-A3B at the final MoE layer (Table~\ref{tab:detailed_domain_routing}). On AIME and HMMT, both math-heavy benchmarks like MATH Level-5, the qualitative pattern of Section~\ref{sec:main_finding} holds: Fragile tokens have low standard-route probability ($\sim 37$), large best-route gaps ($\sim 19$), and very low Top-$1$ rates ($0.6$, $0.7$). GPQA-Diamond shows the same qualitative pattern but less severe: the Fragile fraction is smaller ($5.5$ vs $\sim 10$), the standard-route probability on Fragile tokens ($50.0$) and the Top-$1$ rate ($1.6$) are higher. The best-route gap is comparable in magnitude ($21.2$).

\paragraph{Models.}
We repeat the MATH Level-5 analysis on GPT-OSS-20B, DeepSeek-V2-Lite, and OLMoE-1B-7B at each model's final MoE layer using its native top-$k$ configuration (Table~\ref{tab:cross_model_routing}).
The qualitative pattern of Section~\ref{sec:main_finding} holds across all three: the standard router is well-aligned on Confident tokens and substantially less so on Fragile ones, with Fragile-token best-route gaps of $11.8$, $24.9$, and $20.8$ respectively.
Absolute levels differ across models: OLMoE-1B-7B keeps higher Top-$K$ rates throughout, including on Fragile tokens, while DeepSeek-V2-Lite has very low Top-1 rates even on Confident tokens. 
But in every case the standard route stops tracking route utility once $\bar{p}_{t,\ell}$ leaves the Confident range.
The consistency of this pattern across model families, layers, and domains points to a shared cause that does not depend on which particular router was trained.

\section{A Counterfactual Blind Spot in Standard MoE Training}
\label{sec:blind_spot}

The shared cause lies in how standard top-$k$ MoE training evaluates routing decisions. The language modeling loss measures only the loss of the route that was executed, and load balancing losses depend only on aggregate router-side statistics. Neither places a token-level signal on equal-compute routes that were not executed. We call this gap the counterfactual blind spot. The blind spot predicts the empirical signature observed in Section~\ref{sec:analysis}. Routing decisions are well-trained where their utility is identifiable from the executed loss, and progressively less informative as that identifiability degrades.

\subsection{Formalizing the Blind Spot}
\label{sec:blind_spot_formal}

It is well documented that token-to-expert routing in MoE language models stabilizes early in training and changes little thereafter~\citep{dai2022stablemoe, xue2024openmoe}. The structure of the language-modeling objective explains why. For a token $t$, only the executed route $S_t^{\mathrm{std}}$ enters the forward pass, so differentiating $C_t$ with respect to a router score $s_{tj}$ within a fixed top-$k$ cell yields
\begin{equation}
\frac{\partial C_t}{\partial s_{tj}} \;=\; \mathbf{1}\{j \in S_t^{\mathrm{std}}\}\, g^{\mathrm{std}}_{tj}\!\left(\nabla_{h_t^{\mathrm{std}}} C_t \cdot \bigl(E_j(x_t) - h_t^{\mathrm{std}}\bigr)\right).
\label{eq:exec_grad}
\end{equation}
The indicator is the key term. For $j \in S_t^{\mathrm{std}}$, the gradient refines the scores and gate weights of the experts already in the executed mixture; for $j \notin S_t^{\mathrm{std}}$, it vanishes, and the unselected expert's output $E_j(x_t)$ never enters the computation graph. Two distinct routes $S, S' \subseteq \{1,\dots,N\}$ with $|S|=|S'|=k$ produce different next-token cross-entropies $C_t(S)$ and $C_t(S')$ on the same token, but standard training observes only one of them. 

This is not a gradient bug but the intended sparse computation; the point is that the resulting signal fine-tunes routing within the executed top-$k$ cell while leaving routes in neighboring cells unevaluated. Because the standard router is a single linear projection across the MoE models, there is no separate router-side mechanism that can infer these missing counterfactual losses without an additional training signal. The known stability of trained routing decisions then has a structural reading: once an early routing pattern is established, the gradient signal fine-tunes that pattern but cannot, on its own, propose a better one for tokens where the pattern routes poorly.

\subsection{Load Balancing Does Not Provide Token-Level Counterfactual Signal}
\label{sec:load_balancing}
A natural question is whether the load-balancing losses used in MoE pretraining mitigate this blind spot. 
They can: by discouraging expert collapse, load balancing encourages broader expert coverage and can cause underused experts to be executed more often during training. 
However, this aggregate exploration is not the same as token-level counterfactual supervision. 
We call a load-balancing regularizer aggregate if its value and gradient depend only on router scores, router probabilities, and aggregate routing statistics, rather than on the expert outputs $\{E_j(x_t)\}$ or the cross-entropies $\{C_t(S)\}$ that would result from executing alternative routes on token $t$. 
Such a regularizer may change which experts are explored across a batch, but it does not tell the router which equal-compute alternative would have lowered the loss on a particular hard token.

We make this concrete for the Switch-style load-balancing loss~\citep{fedus2022switch}, a widely-used variant in MoE training. For a batch of $T$ tokens, let $p_t = \mathrm{softmax}(s_t) \in \Delta^{N-1}$ and define
\begin{equation}
f_i = \frac{1}{Tk} \sum_{t=1}^T \mathbf{1}\{i \in S_t^{\mathrm{std}}\}, \qquad \bar{p}_i = \frac{1}{T} \sum_{t=1}^T p_{ti},
\end{equation}
the fraction of routed slots assigned to expert $i$ and its average router probability, with the Switch-style loss given by $R_{\mathrm{lb}} = \lambda N \sum_i f_i \bar{p}_i$. Treating $f$ as fixed, as is standard given the non-differentiability of the routed indicator, only $\bar{p}$ depends differentiably on $s_t$, and the softmax Jacobian gives
\begin{equation}
\nabla_{s_t} R_{\mathrm{lb}} = \frac{\lambda N}{T} \big( \mathrm{diag}(p_t) - p_t p_t^\top \big) f.
\label{eq:lb_grad}
\end{equation}
This gradient depends only on $f$ and $p_t$ but not on $\{E_j(x_t)\}$ or $\{C_t(S)\}$, confirming that the Switch-style loss is aggregate in the sense defined above. 
The same property holds for the other variants used in current open-weight MoE models: the Importance and Load losses of \citet{shazeer2017outrageously}, the GShard auxiliary loss~\citep{lepikhin2020gshard}, the device-level balance terms of \citet{liu2024deepseek2}, and the auxiliary-loss-free bias update of \citet{wang2024auxiliary} adopted in \citet{liu2024deepseek3}.
These mechanisms are important for preventing expert collapse and improving aggregate coverage, but they do not directly supply the missing counterfactual route-loss signal. 
Closing the blind spot studied here would require a signal that compares the loss of the executed route with the losses of alternative equal-compute routes for the same token.


\section{Probing the Routing Blind Spot with a Minimal Update}
\label{sec:experiments}

Section~\ref{sec:blind_spot} left the two mechanisms underlying high-loss tokens mixed: intrinsic difficulty and routing failure both contribute, and the executed-route loss alone cannot tell them apart. We now partially separate them by intervening only on routing inside the frozen model. If a minimal router-only update is sufficient to shift downstream pass@$K$, then at least part of the high loss on hard tokens is reachable by re-routing alone rather than requiring expert re-training. Concretely, we update only the final-layer router while leaving every expert and every other router frozen. At evaluation time, the model uses ordinary top-$k$ routing (with updated parameters) with no test-time search.

\subsection{Expert Preference Optimization}
\label{sec:method}

\paragraph{Method.}
To this end, we propose Expert Preference Optimization (EPO), a minimal router-only update that intervenes only on routing inside the frozen model.
At each training step, we define hard tokens as tokens whose current next-token cross-entropy under the trainable router $\pi_\theta$ exceeds $\tau=0.1$. 
For each such token $t$, let $r^-_t$ be the top-$k$ route currently selected by $\pi_\theta$. We sample $G$ alternative top-$k$ routes from $\pi_\theta$ via Gumbel-top-$k$, all with the same number of routed experts as $r^-_t$. For each sample we recompute the gate weights, execute that route at the final MoE layer, and measure the resulting next-token cross-entropy with all other parameters fixed. If the lowest-CE sample improves over $r^-_t$, we use it as the chosen route $r^+_t$; otherwise the token contributes no gradient at this step.

Let $\pi_{\text{ref}}$ denote the frozen standard router and $\pi_\theta$ the trainable router, initialized from $\pi_{\text{ref}}$. We treat each route $r$ as a sequence of experts and use the factorized log-probability $\log \pi(r \mid x_t) = \sum_{e \in r} \log \pi(e \mid x_t)$, where $\pi(e \mid x_t)$ is the router softmax probability assigned to expert $e$. We then train $\pi_\theta$ with a CE-gap-weighted DPO-style objective~\citep{amini2024odpo, zhou2024wpo}:
\begin{equation}
\ell_t = -\Delta_t \log \sigma\left(\beta \log \frac{\pi_\theta(r^+_t \mid x_t)}{\pi_{\text{ref}}(r^+_t \mid x_t)} - \beta \log \frac{\pi_\theta(r^-_t \mid x_t)}{\pi_{\text{ref}}(r^-_t \mid x_t)}\right),
\end{equation}
where $\Delta_t = \max(0, C(r^-_t) - C(r^+_t))$ is the observed CE-gap weight.

Since $r^+_t$ and $r^-_t$ have the same cardinality, shared experts cancel in the route log-probability difference and the softmax normalizers cancel as well. The update therefore acts only on experts that distinguish $r^+_t$ from $r^-_t$: chosen-only experts are pushed up, rejected-only experts are pushed down. EPO 
modifies the final-layer routing weights for the experts that separate a lower-loss route from the route currently chosen by the trainable router, without updating any expert weights.

\paragraph{Setup.}
We evaluate EPO on Qwen3-30B-A3B~\citep{yang2025qwen3} and GPT-OSS-20B~\citep{agarwal2025gpt}. We construct router-update data from MATH Level-5~\citep{hendrycks2021measuring}: for each problem we sample four solutions from the standard model and keep problems with at least one verified-correct solution, yielding approximately 2K verified trajectories. The hard-token filter described above is applied at each training step on these trajectories. We evaluate on AIME 2024+2025 and HMMT 2025 with the same decoding settings and number of samples per problem for both standard and EPO models.

\subsection{Pass@K Shifts under a Router-Only Probe}
\label{sec:epo_results}

\begin{figure}[t]
    \centering
    \begin{subfigure}[b]{0.47\linewidth}
        \centering
        \includegraphics[width=\linewidth]{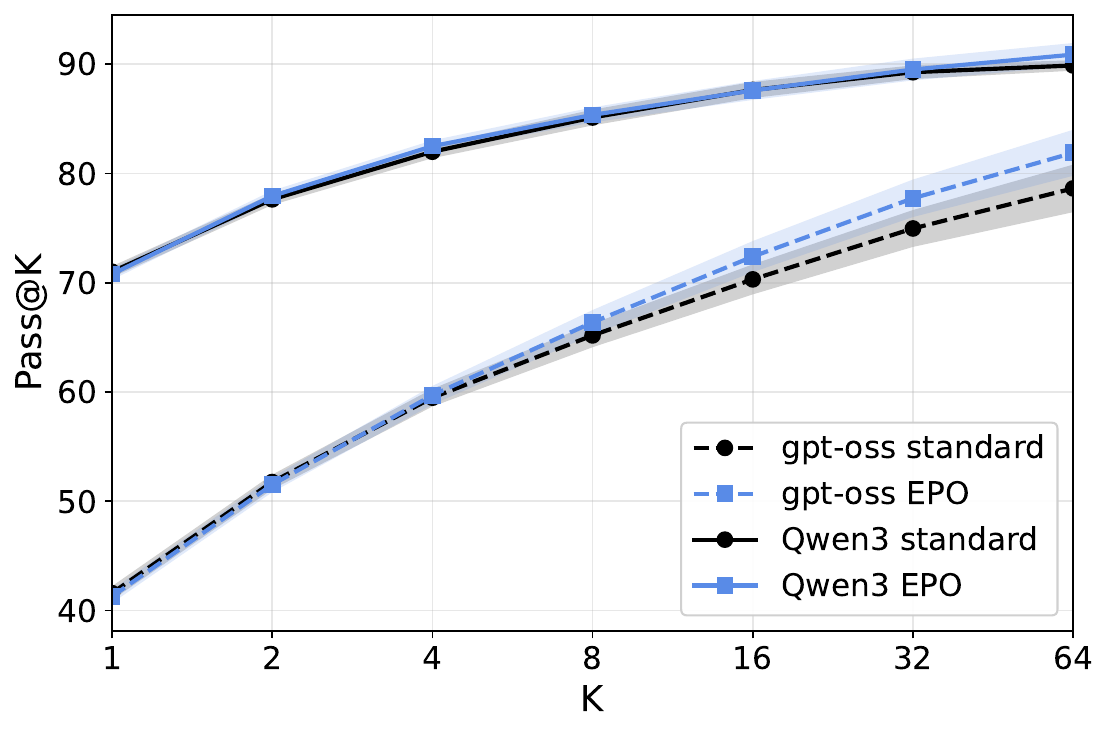}
        \caption{AIME 2024 + 2025}
        \label{fig:passk_aime}
    \end{subfigure}
    \hfill
    \begin{subfigure}[b]{0.47\linewidth}
        \centering
        \includegraphics[width=\linewidth]{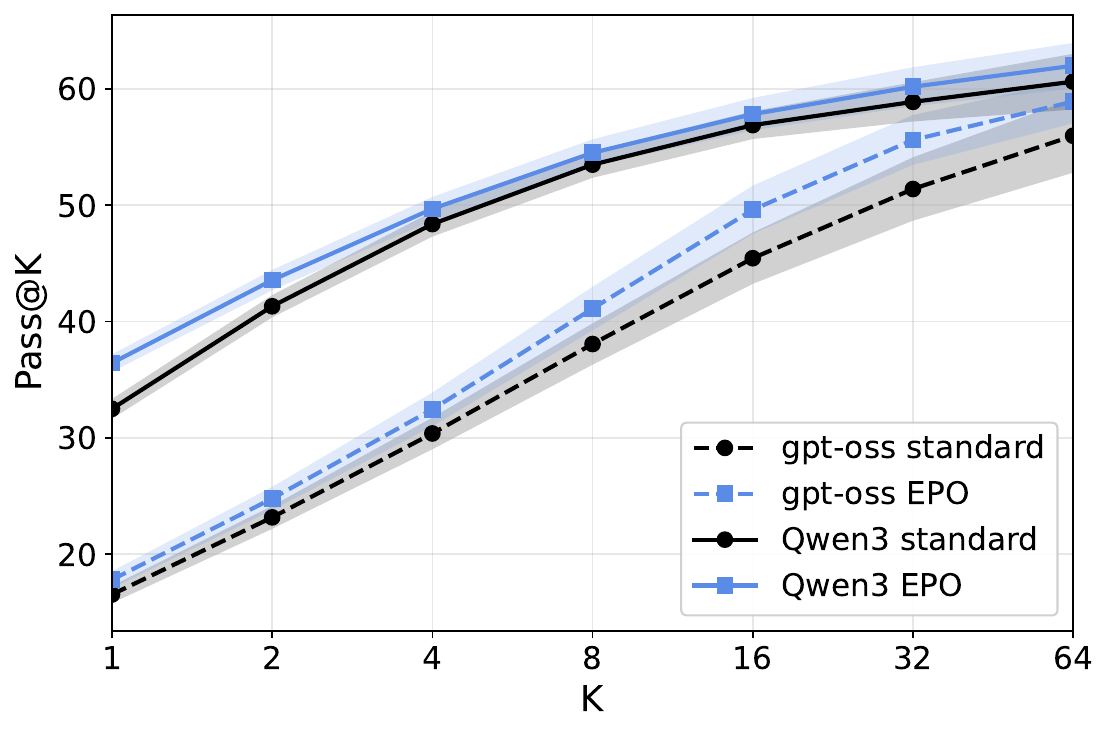}
        \caption{HMMT 2025}
        \label{fig:passk_hmmt}
    \end{subfigure}
    \caption{
        \textbf{Pass@$K$ curves on AIME 2024+2025 and HMMT 2025.} For each problem, we pool $n=160$ completions and compute pass@$K$ with the standard combinatorial estimator $\widehat{q}_p(K)=1-\binom{n-c_p}{K}/\binom{n}{K}$, where $c_p$ is the number of correct completions. Scores are averaged over problems. Shaded bands show 95\% bootstrap confidence intervals; see Appendix~\ref{app:Experiments} for details.
    }
    \label{fig:passk}
    \vspace{-0.3cm}
\end{figure}

\paragraph{Shifts under the router-only update.}
As illustrated in Figure~\ref{fig:passk}, the standard and EPO curves are close on both benchmarks and both models, but the EPO curves sit above the standard curves at most reported $K$, and the $95\%$ bootstrap bands separate from the standard over part of the reported range. The shifts are small and we read them as a minimal existence check rather than as a routing improvement method: a router-only update on the order of $0.001\%$ of model parameters, with no expert or attention block modified, is sufficient to move pass@$K$ above the standard curve under bootstrap uncertainty. We interpret this as evidence that part of the high loss on hard tokens is reachable inside the frozen model by re-routing alone, rather than requiring expert re-training.

\paragraph{Where does the shift come from?}
Table~\ref{tab:trajectory_analysis} decomposes EPO's effect by separately varying the router used for ordinary top-$k$ selection and the trajectory being analyzed. `A' is the standard baseline. `B' holds the generated trajectory fixed and evaluates the EPO router, isolating how the updated router changes route selection on the standard-model trajectory. `C' uses the EPO router end-to-end, capturing both changed routing and the resulting change in generated trajectories. 
First, EPO shifts the token distribution toward easier bins: the Confident fraction increases from $73.2$ in A to $74.5$ in B and $75.8$ in C, while the Fragile fraction decreases from $9.8$ to $9.6$ and then $8.2$. 
Second, the route-rank gains occur mainly outside the Confident bin. 
On Ambiguous tokens, A $\to$ B raises Top-1 from $1.2$ to $2.4$, with corresponding gains in Top-5 and Top-10. 
On Fragile tokens, Top-1 rises from $0.6$ in A to $1.9$ in B and $2.4$ in C, again with Top-5 and Top-10 improving as well.

\paragraph{Interpreting the Confident-bin rank drop.}
The decline of Top-$K$ rates on Confident tokens should be interpreted differently. 
In this bin, route choice has little measurable effect: $p_{\mathrm{std}}$ remains around $99.5$, $p_{\mathrm{best}}$ around $99.8$, and the Gap stays at only $0.3$ percentage points across A, B, and C. 
Thus the lower Confident-bin rank rates do not indicate a meaningful loss in next-token probability. 
Rather, EPO appears to reduce an easy-token-biased rank alignment of the original router, while improving rank alignment on Ambiguous and Fragile tokens where different equal-compute routes lead to substantially different realized-token probabilities. 
We therefore interpret Table~\ref{tab:trajectory_analysis} as evidence that the router-only update changes route selection in the regimes where route choice matters, while leaving already-confident predictions essentially unchanged at the probability level.

\begin{table}[t]
\centering
\caption{\textbf{EPO's effect on routing quality.} A: standard trajectory with the pre-trained router. B: standard trajectory with EPO-updated router. C: EPO trajectory with EPO-updated router.}
\label{tab:trajectory_analysis}
\footnotesize
\setlength{\tabcolsep}{4pt}
\begin{tabular}{lR{1.08cm}R{1.08cm}R{1.08cm}R{1.08cm}R{1.08cm}R{1.08cm}R{1.08cm}R{1.08cm}R{1.08cm}}
\toprule
& \multicolumn{3}{c}{Confident} & \multicolumn{3}{c}{Ambiguous} & \multicolumn{3}{c}{Fragile} \\
\cmidrule(lr){2-4} \cmidrule(lr){5-7} \cmidrule(lr){8-10}
& \multicolumn{1}{c}{A} & \multicolumn{1}{c}{B} & \multicolumn{1}{c}{C}& \multicolumn{1}{c}{A} & \multicolumn{1}{c}{B} & \multicolumn{1}{c}{C}& \multicolumn{1}{c}{A} & \multicolumn{1}{c}{B} & \multicolumn{1}{c}{C} \\
\midrule
Tokens (\%) & 73.2 & 74.5 & 75.8 & 17.0 & 16.0 & 16.0 & 9.8 & 9.6 & 8.2 \\
Top-1 (\%) & 42.7 & 27.0 & 26.3 & 1.2 & 2.4 & 1.9 & 0.6 & 1.9 & 2.4 \\
Top-5 (\%) & 63.4 & 43.5 & 43.3 & 9.4 & 16.8 & 16.2 & 11.2 & 17.9 & 20.7 \\
Top-10 (\%) & 76.1 & 60.6 & 61.1 & 25.5 & 35.7 & 35.7 & 37.2 & 39.7 & 44.5 \\
$p_{\mathrm{std}}$ (\%) & 99.5 & 99.4 & 99.5 & 77.2 & 77.7 & 77.9 & 37.5 & 35.0 & 39.0 \\
$p_{\mathrm{best}}$ (\%) & 99.8 & 99.8 & 99.8 & 88.1 & 89.2 & 89.3 & 56.6 & 56.8 & 59.5 \\
Gap (pp)& 0.3 & 0.3 & 0.3 & 10.8 & 11.5 & 11.3 & 19.2 & 21.9 & 20.5 \\
\bottomrule
\end{tabular}
\vspace{-0.3cm}
\end{table}

\section{Related Work}
\label{sec:related_work}

\paragraph{MoE training.}
A large body of work makes sparse routing scalable, balanced, and stable during 
pretraining. Auxiliary load-balancing losses prevent expert collapse 
\citep{shazeer2017outrageously, lepikhin2020gshard, fedus2022switch}, 
auxiliary-loss-free balancing replaces these with bias updates from observed 
expert load \citep{wang2024auxiliary}, and the router $z$-loss controls router 
logit magnitudes for stability \citep{zoph2022st}. A separate line modifies the 
routing rule itself \citep{lewis2021base, zhou2022mixture, puigcerver2023sparse, 
wang2024remoe}. After pretraining, MoE-specific methods adapt task-relevant 
experts \citep{wang2024let}, use route exploration as an RL signal for the 
router \citep{kim2025exploration}, constrain routing drift during fine-tuning 
\citep{kim2025defending}, or incorporate expert selection into RL post-training 
\citep{ko2026moe, ma2026balancing}. These works improve how routing is learned 
or adapted, but none directly evaluates whether the routes selected by a trained 
top-$k$ router are counterfactually good.

\paragraph{Token-level analysis of reasoning.}
Reasoning success is determined disproportionately by a small subset of tokens: 
pivotal tokens have outsized effect on final-answer correctness 
\citep{abdin2024phi}, intervening on critical tokens identified by token-level 
contrastive estimation substantially changes reasoning outcomes 
\citep{lin2024critical}, and high-entropy minority tokens drive most of the 
policy improvement during RL training \citep{wang2025beyond}. 
Our Fragile bin is not identical to these prior definitions of pivotal, critical, or high-entropy tokens. 
It is a route-conditional diagnostic: a token is Fragile when sampled equal-compute routes assign low average probability to the realized continuation. 
The connection is therefore conceptual rather than operational. 
Both lines of evidence suggest that aggregate token averages can hide failures concentrated on a small number of uncertain positions, where local decisions may matter disproportionately.

\paragraph{Analyzing routing in trained MoE.}
Prior analyses of trained MoE models document expert specialization, routing 
saturation, and domain or vocabulary structure 
\citep{xue2024openmoe, muennighoff2024olmoe}. These ask what routing has 
learned, framed as which expert receives which token. We ask an orthogonal 
question: given the route the router selected, was it a good choice relative 
to other equal-compute routes inside the same frozen model. A router can route 
consistent token types to consistent experts and still select a suboptimal 
expert set on the tokens that matter most. A separate line intervenes on 
routing at inference time, by sampling from uncertain expert choices 
\citep{chen2026certain}, pruning or reinforcing experts without retraining 
\citep{chen2025ban}, or optimizing expert mixtures per input 
\citep{li2025c3po, li2025r2}. These also treat the default route as not 
necessarily optimal, with \citet{chen2026certain} in particular tying the 
intervention to token-level uncertainty. Our analysis is upstream: we measure 
whether the default route is counterfactually good in the first place, under 
the frozen model and without test-time search.
\section{Conclusion}
\label{sec:conclusion}
We analyzed routing quality in trained MoE language models by comparing each 
standard top-$k$ route against sampled equal-compute alternatives under the 
frozen model. The standard router is well-aligned with route utility on 
confident tokens but uninformative on the fragile tokens that drive hard 
reasoning, and this token-conditional pattern holds across four open-weight 
MoE families. The pattern is consistent with a structural property of 
standard top-$k$ training: the language modeling loss evaluates only the 
executed route, and load balancing depends only on aggregate routing 
statistics, so neither places a token-level signal on the equal-compute 
routes that were not selected. A minimal router-only update on hard tokens is sufficient to shift downstream pass@$K$ on mathematical reasoning tasks, indicating that the failure reflects misallocation reachable by the router alone rather than a pure capacity limit. We take these results to suggest that aggregate 
routing metrics are not sufficient to detect failures of routing quality, 
and that routing on hard tokens warrants attention as a target of MoE 
training rather than as an implementation detail of sparsity.

\paragraph{Limitations.}
We analyze each layer's routing independently, holding the other layers 
fixed; whether per-layer failures compound or cancel when routing is 
modified jointly across layers is a separate question we do not address. 
Our analysis also restricts attention to verified-correct trajectories, 
where the realized next token already lies on a successful path; we use 
its probability under route intervention as a proxy for routing quality 
on hard tokens, but this proxy may overstate the room for improvement on 
tokens where the realized token was itself a fortunate sample.

\paragraph{From diagnosis to prevention.}
Our analysis is performed on already-trained MoE models, but the failure it documents is a property of how these models are trained: standard top-$k$ pretraining provides no token-level signal on routes that were not executed, and the misalignment we observe at fragile tokens is the predictable consequence. 
The natural follow-up is therefore not to refine the post-hoc router-only update, but to address the cause at training time. 
Concretely, this means designing a pretraining objective that places token-level signal on equal-compute alternative routes, for example by lightweight evaluation of sampled alternative routes alongside the executed one and incorporating their cross-entropies into the loss. 
Whether such a counterfactual-aware objective can be implemented at pretraining scale without sacrificing the efficiency advantages of sparse computation, and whether routers trained under it avoid the difficulty-conditioned failure documented here in the first place, are the questions our analysis points to.

\bibliography{neurips_2026}  
\bibliographystyle{apalike}  

\begin{appendix}

\section{Model Details}
\label{app:LLMs}
We analyze four open-weight Mixture-of-Experts language models. For 
each model we use the instruction-tuned variant publicly released on 
Hugging Face (Qwen3-30B-A3B-Instruct-2507, gpt-oss-20b 
with reasoning effort set to low, DeepSeek-V2-Lite-Chat, 
and OLMoE-1B-7B-0125-Instruct), evaluated in its native 
top-$k$ configuration without any additional fine-tuning except as 
described in Section~\ref{sec:method}.

\section{Hyperparameter Details}
\label{app:Hyperparameter}
This appendix documents the hyperparameters used in the counterfactual 
routing analysis (Section~\ref{sec:analysis}) and in the router-only 
update (Section~\ref{sec:experiments}). Table~\ref{tab:analysis_hyperparams} 
lists the analysis hyperparameters, which are shared across all four MoE 
models we study, with model-specific values given for top-$k$ and the 
target layer. The pool size of 32 refers to the size of the candidate 
expert pool over which Gumbel-top-$k$ samples are drawn: at each token 
we restrict alternative routes to the 32 experts with the highest router 
scores, then perturb their scores with Gumbel noise. 
Table~\ref{tab:epo_hyperparams} lists the router-only update 
hyperparameters separately for Qwen3-30B-A3B and GPT-OSS-20B.

\begin{table}[!ht]
\centering
\caption{Hyperparameters for the counterfactual routing analysis (Section~\ref{sec:analysis}).}
\label{tab:analysis_hyperparams}
\small
\begin{tabular}{lll}
\toprule
Parameter & Value & Description \\
\midrule
$G$ & 32 & Number of alternative routes per token \\
Pool size & 32 & Candidate expert pool size \\
$k$ & 8 / 4 / 6 / 8 & top-$k$ for Qwen3 / GPT-OSS / DSv2 / OLMoE \\
Noise scale & 1.0 & Gumbel scale parameter \\
Target layer & L47 / L23 / L26 / L15 & Final MoE layer for Qwen3 / GPT-OSS / DSv2 / OLMoE \\
Random seed & 42 & For subsample reproducibility \\
\texttt{dtype} & \texttt{bfloat16} & Inference precision \\
Number of problems & 100 / 10 / 10 / 100 & \makecell[l]{Randomly sampled per benchmark \\ (MATH-L5 / AIME / HMMT / GPQA-Diamond)} \\
\bottomrule
\end{tabular}
\end{table}

\begin{table}[!ht]
\centering
\caption{Hyperparameters for the router-only update (Section~\ref{sec:experiments}).}
\label{tab:epo_hyperparams}
\small
\begin{tabular}{lll}
\toprule
Parameter & Qwen3-30B-A3B & GPT-OSS-20B \\
\midrule
Target layer & L47 & L23 \\
Optimizer & AdamW & AdamW \\
Learning rate & $3 \times 10^{-4}$ & $3 \times 10^{-4}$ \\
Weight decay & 0.01 & 0.01 \\
$\beta_{\text{DPO}}$ & 0.1 & 0.1 \\
Epochs & 1 & 1 \\
$G$ & 32 & 16 \\
Pool size & 32 & 16 \\
$k$ & 8 (native) & 4 (native) \\
Noise scale & 1.0 & 1.0 \\
Hard-token threshold $\tau$ & 0.1 & 0.1 \\
Batch size & 16 & 16 \\
Number of problems & 2{,}269 & 2{,}162 \\
Gradient norm clip & 1.0 & 1.0 \\
Random seed & 42 & 42 \\
\bottomrule
\end{tabular}
\end{table}

\section{Experimental Details}
\label{app:Experiments}
This appendix documents the data construction and evaluation procedures 
used in the routing-quality analysis (Section~\ref{sec:analysis}) and 
in the router-only update (Section~\ref{sec:experiments}). For both, we 
generate self-sampled solutions on MATH Level-5 problems, retain 
verified-correct trajectories using exact-answer matching after 
normalization, and restrict attention to assistant-response tokens. The 
router-only update uses 2,269 verified trajectories for Qwen3-30B-A3B 
and 2,162 for GPT-OSS-20B. At each training step, the hard-token filter selects tokens whose current cross-entropy under the trainable router exceeds the threshold $\tau$ in Table~\ref{tab:epo_hyperparams}, and only these tokens contribute gradient at that step.

Pass@$K$ evaluation uses AIME 2024+2025 (60 problems) and HMMT February 
2025 (30 problems). For each problem we generate $n = 160$ completions 
(32 samples per run, pooled across 5 runs with different random seeds), 
and compute the per-problem pass@$K$ with the unbiased combinatorial 
estimator $\hat{q}_p(K) = 1 - \binom{n - c_p}{K} / \binom{n}{K}$, where 
$c_p$ is the number of correct completions. The reported pass@$K$ is 
the mean of $\hat{q}_p(K)$ across problems. To quantify the uncertainty 
of this mean, we apply parametric bootstrap (2,000 resamples) over the 
binomial sampling of $c_p$ at $n = 160$: for each problem we draw 
$c_p^{(b)} \sim \text{Binomial}(n, c_p / n)$, recompute 
$\hat{q}_p^{(b)}(K)$, and average across problems to obtain a bootstrap 
replicate of the mean. The 95\% confidence intervals shown as shaded 
bands in Figure~\ref{fig:passk} are the 2.5th and 97.5th percentiles 
of the 2,000 bootstrap replicates. Decoding settings are listed in 
Table~\ref{tab:decoding}.

\begin{table}[!ht]
\centering
\caption{Decoding settings for AIME and HMMT pass@$K$ evaluation.}
\label{tab:decoding}
\small
\begin{tabular}{lll}
\toprule
Setting & Qwen3-30B-A3B & GPT-OSS-20B \\
\midrule
Temperature & 0.6 & 1.0 \\
\texttt{top\_p} & 0.95 & 0.95 \\
\texttt{top\_k} & --- & 20 \\
Samples per problem & \makecell[l]{32 $\times$ 5 runs \\ ($n = 160$)} & \makecell[l]{32 $\times$ 5 runs \\ ($n = 160$)} \\
Max tokens & 8192 & 8192 \\
Reasoning effort & --- & \texttt{low} \\
\bottomrule
\end{tabular}
\end{table}

\paragraph{Compute.}
All experiments were conducted on NVIDIA RTX A6000 GPUs (48 GB). 
Qwen3-30B-A3B uses two A6000 GPUs and the other models 
(GPT-OSS-20B, DeepSeek-V2-Lite, OLMoE-1B-7B) each use a single A6000 
GPU. Because the update procedure computes and stores gradients only 
for the router parameters, the GPU memory footprint is essentially 
that of the forward pass plus the model weights. Approximate wall-clock 
times for each experiment are listed in Table~\ref{tab:compute_time}. 
The longer runtime for Qwen3-30B-A3B reflects inter-GPU communication 
overhead from sharding the model across two devices; running the same 
procedure on a single larger-memory GPU (e.g., A100 80 GB) would likely 
reduce wall-clock time substantially.

\begin{table}[!ht]
\centering
\caption{Approximate wall-clock time per experiment.}
\label{tab:compute_time}
\small
\begin{tabular}{lcc}
\toprule
Model & Routing-quality analysis & Router-only update \\
\midrule
Qwen3-30B-A3B & 2 hours & 5 hours \\
GPT-OSS-20B & 17 minutes & 20 minutes \\
DeepSeek-V2-Lite & 19 minutes & --- \\
OLMoE-1B-7B & 18 minutes & --- \\
\bottomrule
\end{tabular}
\end{table}

\paragraph{Software environment.}
Inference uses vLLM 0.10.1 with HuggingFace Transformers 4.57.6 on 
PyTorch 2.7.1 (Python 3.11). 
Pass@$K$ evaluation uses lighteval 0.13.0 
and inspect\_ai 0.3.202, with answer verification by exact-match 
comparison after standard normalization.

\end{appendix}


\end{document}